\documentclass[letterpaper, 10 pt, conference]{ieeeconf}  

\IEEEoverridecommandlockouts            %
\usepackage{graphics} %
\usepackage{graphicx} %
\usepackage{amsmath} %
\usepackage{amssymb}  %
\usepackage{mathtools}
\usepackage{algpseudocode}
\usepackage{algorithm}
\usepackage{tikz}
\usetikzlibrary{calc}

\usepackage{subcaption}

\graphicspath{ {../figs/} }

\title{\LARGE \bf Learning to Navigate: Exploiting Deep Networks to Inform
Sample-Based Planning During Vision-Based Navigation}

\author{Justin S. Smith$^{1}$, Jin-Ha Hwang$^{1}$,  Fu-Jen Chu$^{1}$, 
    and Patricio A. Vela$^{1}$%
\thanks{*This work supported in part by NSF Awards \#1400256 and \#1605228.}%
\thanks{$^{1}$J.S. Smith, J. Hwang, F.J. Chu, and P.A. Vela are with the School of
Electrical and Computer Engineering and the Institute for Robotics and
Intelligent Machines, Georgia Institute of Technology, Atlanta, GA 30308,
USA.
{\tt\small \{jssmith, jhwang44, fujenchu, pvela\}@gatech.edu}}%
}

\begin{document}

\maketitle
\thispagestyle{empty}
\pagestyle{empty}

\begin{abstract}
Recent applications of deep learning to navigation have generated end-to-end
navigation solutions whereby visual sensor input is mapped to control
signals or to motion primitives.  The resulting visual navigation
strategies work very well at collision avoidance and have performance that
matches traditional reactive navigation algorithms while operating in
real-time.  It is accepted that these solutions cannot provide the same
level of performance as a global planner.  However, it is less clear how
such end-to-end systems should be integrated into a full navigation pipeline.
We evaluate the typical end-to-end solution within a full navigation pipeline
in order to expose its weaknesses.  Doing so illuminates how to better
integrate deep learning methods into the navigation pipeline.  In particular,
we show that they are an efficient means to provide informed samples for
sample-based planners.  Controlled simulations with comparison against
traditional planners show that the number of samples can be reduced by an
order of magnitude while preserving navigation performance.  Implementation
on a mobile robot matches the simulated performance outcomes.
\end{abstract}

\section{Introduction}

Contemporary machine learning (ML) algorithms demonstrate the potential to
provide end-to-end learning such that navigation tasks can be replaced with
black box image-to-control or image-to-action mappings 
\cite{RoEtAl_ICRA[2013],SaLe_RSS[2017]}.  
This data-driven approach is inline with-- but on a path to supercede--
traditional (manually engineered) reactive navigation strategies. The idea
behind ML methods is that the underlying learner, today usually a deep
network, implicitly learns structural models of the world within its
internal representation.  These implicit structural models are then encoded
in the feature vector output of the network and support effective decision
making at the output end, following the incorporation of a shallow learner.

The traditional deliberative, or global, navigation pipeline is heavily
model based (in the sense that it requires models of the robot and the
world) and is at odds with end-to-end machine learning approaches.  Several
ML-navigation based papers further conclude that it is best to fuse their
algorithms with higher level planners to overcome the short-term, greedy
nature of ML-navigation \cite{SaLe_RSS[2017],PfEtAl_ICRA[2017]}.  
One structural issue with doing so is that typical end-to-end systems are
often not meant to consider higher level planning actions, thus there is no
obvious procedural strategy for achieving the desired recommendation. An
exception being ML-strategies that explicitly incorporate the task
\cite{PfEtAl_ICRA[2017]}, or that learn motion primitives
\cite{ZhEtAl_ICRA[2016]}.  The latter can be concatenated with a higher
level planner. Additional work is needed though to align the motion
primitive control decisions with the global goal.

In this paper, we explore the standard ML pipeline used for end-to-end
navigation and demonstrate that it is not well suited to incorporation into
the traditional, closed-loop, goal-directed navigation pipeline.
Furthermore, since traditional global navigation approaches involve 
exhaustive or highly sampled search, their performance relative to
ML-methods should be considered as upper bounds.  Taking a systems
view of vision-based navigation, we propose sensible modifications to
ML-based navigation methods so that they are more compatible with the
traditional goal-directed navigation stack.  
Importantly, the input/output function learnt must be altered to
consider down-stream use of the ML output.  Implementing the modified
ML-driven approaches in controlled simulations we show that a
planning-aware ML-based approach to navigation can indeed provide
comparable performance to more global or exhaustive approaches while
reducing the trajectory search space. The ML-based module effectively
acts as a data-driven sampling heuristic for minimizing the trajectory
search and providing presumably high quality navigation directions.  The
resulting planner leverages learning where learning excels, and relies
on model-based strategies where they excel.

\subsection{Related Work}

Due to the extensive literature and history of research efforts on planning,
this review of related work focuses primarily on contemporary learning-based
approaches to navigation.  It is presumed that the reader has some
familiarity with planning in general \cite{LaValle[2006]}.

Collision-free navigation requires processing sensor measurements to
understand or infer the local scene geometry, then making a navigation
decision based on both the scene geometry and the global goal points.  For
mobile robots with monocular cameras, depth information must be inferred
from the sensing time signal. As an input/ouput mapping, ML has shown
success in recovering good estimates of scene structure for navigation
purposes \cite{BiDuKr_ICRA[2015],LaEtAl_IROS[2016],MiSaNg_ICML[2005]}. 
The ML block generates missing information for decision making.

\tikzstyle{block} = [draw, fill=blue!20, rectangle, rounded corners,
                     minimum height=2em, minimum width=4em]
\tikzstyle{optBlock} = [draw, fill=blue!05, rectangle, rounded corners,
                     minimum height=2em, minimum width=4em, dashed]
\tikzstyle{slowBlock} = [draw, fill=red!05, rectangle, rounded corners,
                     minimum height=2em, minimum width=4em]
\tikzstyle{newtip} = [->, thick]

\begin{figure*}[t!]
\small
\vspace{.15cm}
\hfill
\begin{tikzpicture}[auto, node distance=2cm,>=latex, scale=0.2]
  \node[anchor=north west, block] (World) at (0in,0in) {Evolve};
  \node[block, anchor=west] (Sense) at ($(World.east) + (1.5cm,0cm)$)
    {Sense};
  \node[block, anchor=west] (Plan) at ($(Sense.east) + (1.5cm,0cm)$) {React};
  \node[block, anchor=north] (Control) at ($(Sense.south) + (0cm,-2.0cm)$)
    {Control};
  \draw[newtip] (World) -- (Sense);
  \draw[newtip] (Sense) -- (Plan);
  \draw[newtip] (Plan.south) |- (Control);
  \draw[newtip] (Control) -| (World.south);
  \node at (1cm, -8cm) {(a)};
\end{tikzpicture}
\hfill
\begin{tikzpicture}[auto, node distance=2cm,>=latex, scale=0.2]
  \node[block,anchor=north west] (World) at (0in,0in) {Evolve};
  \node[block, anchor=west] (Sense) at ($(World.east) + (1.5cm,0cm)$)
    {Sense};
  \node[slowBlock, anchor=west] (Fuse) at ($(Sense.east) + (1.5cm,0cm)$)
    {Fuse};
  \node[slowBlock, anchor=west] (Plan) at ($(Fuse.east) + (1.5cm,0cm)$) {Plan};
  \node[block, anchor=north] (Control) at ($(Sense.south) + (0cm,-2.0cm)$)
    {Control};
  \draw[newtip] (World) -- (Sense);
  \draw[newtip] (Sense) -- (Fuse);
  \draw[newtip] (Fuse) -- (Plan);
  \draw[newtip] (Plan.south) |- (Control);
  \draw[newtip] (Control) -| (World.south);
  \node at (1cm, -8cm) {(b)};
\end{tikzpicture}
\hfill
\begin{tikzpicture}[auto, node distance=2cm,>=latex, scale=0.2]
  \node[block, anchor=north west] (World) at (0in, 0in) {Evolve};
  \node[block, anchor=west] (Sense) at ($(World.east) + (1.5cm,0cm)$)
    {Sense};
  \node[slowBlock, anchor=west] (Fuse) at ($(Sense.east) + (3.0cm,0cm)$)
    {Fuse};
  \node[slowBlock, anchor=west] (Plan) at ($(Fuse.east) + (1.5cm,0cm)$)
    {Plan};
  \node[block, anchor=north] (Control) at ($(Sense.south) + (0cm,-2.0cm)$)
    {Control};
  \node[block, anchor=north] (LocPlan) at ($(Fuse.south) + (0cm,-2.0cm)$)
    {React};
  \draw[newtip] (World) -- (Sense);
  \draw[newtip] (Sense) -- (Fuse);
  \draw[newtip] (Fuse) -- (Plan);
  \draw[newtip] (Plan.south) |- (LocPlan.east);
  \draw[newtip] (Sense.south) |- ($(Sense.south) + (0.5,-0.5cm)$)
    -| (LocPlan.north);
  \draw[newtip] (LocPlan) -- (Control);
  \draw[newtip] (Control) -| (World.south);
  \node at (1cm, -8cm) {(c)};
\end{tikzpicture}
\hfill

\vspace*{-0.05in}
\caption{Block diagram of (a) local/reactive; (b) global/deliberative; and
(c) mixed navigation pipelines. \label{planBlocks}}
\vspace*{-0.15in}
\end{figure*}
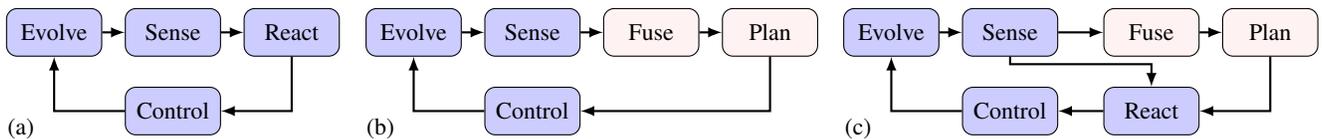

Others propose an  end-to-end process, whereby image to control decisions
are learnt instead \cite{RoEtAl_ICRA[2013],SaLe_RSS[2017]}. 
In the cases where expert demonstration is employed, it is not clear that
the expert's motives for the control decisions necessarily translate to
desirable navigation choices given the mobile robot's goals.  When the
control objective is clear, as in stabilization, then the image to control
mapping is effectively injective, thereby reducing mismatch between implicit
information contained in the training set and the explicit goal 
\cite{VaEtAl_IROS[2016]}. %
Another issue that arises with many end-to-end methods is that the
underlying strategy is functionally a regressor \cite{TaLiLi_IJARS[2017]}.
For navigation, which is non-convex due to the topological splits that occur
in trajectory space, regression may not be the best choice.
Similar problems may arise for methods that learn motion primitives for
specific scenarios \cite{ZhEtAl_ICRA[2016]}, as a particular motion
primitive may be biased towards one topological instance of a reaction
from a set of potential reactions.  The bias may undermine the global
objective.

The work in \cite{PfEtAl_ICRA[2017]} is one of the the first to consider
the goal state within end-to-end training, which involves laser scans and
relative target waypoints as input and steering commands as output.
However, their system still requires human intervention to correct for local
minima since it is not linked to higher level planning.  It is not clear
how successful the system is at autonomous operation relative to the global
objective since baseline comparison against standard methods is not
performed.
An intermediate family of ML-navigation methods convert the image into
actionable information (or affordance states), such as local structural
information useful for driving along roadways \cite{ChEtAl_ICCV[2015]}
(e.g., distance to vehicles, lane locations, lane marker relative offsets,
etc.).  Another example learns to identify forest trails for navigation
guidance \cite{GiEtAl_RAL[2016]}.  The output states are used to then
generate a navigation decision.  The target application is usually a
restricted form of navigation.

When input imagery is too novel or reflects extreme imaging conditions,
then ML-models may fail at the trained task.  Detecting failure, or the
potential of failure, can itself be cast into a learning framework to
thereby supervise the online end-to-end ML controller
\cite{DaEtAl_IROS[2016],SaKuHe_ICRA[2017]}.  In \cite{RiRo_RSS[2017]},
the supervisor is an autoencoder whose reconstruction should match the
scene.  Failure to reconstruct the input image indicates an
out-of-sample situation requiring a fall-back plan.

Within the area of Reinforcement Learning (RL), it is common
to learn image-to-action mappings, usually in the context of video games
\cite{GoEtAl_NIPS[2014],KeEtAl_CIG[2016],LiEtAl_ICAAMS[2016],%
MoEtAl_CVPR[2016],WuEtAl_NIPS[2015]}.
By virtue of an existing reward function, the methods learn task-oriented
behavior.  Though the imagery may be quite diverse, the underlying task
domain is usually limited in scope relative to the general task of
navigation, and the approaches seek a slightly different purpose
from the one stated here.  Though
\cite{MiEtAl_ICLR[2017],ZhEtAl_ICRA[2017]} concern navigation and even
consider specific task-based objectives, the aim of the research
is to demonstrate implicit learning of the environment from imagery based on
repeated trials within the environment.  Changing the environment requires
further rounds of training.

\subsection{Robotic Navigation Systems}

Figure \ref{planBlocks} depict three varieties of navigation pipelines. The
majority of ML-based navigation methods operate at the reactive level,
Figure \ref{planBlocks}a, whereas some of the task-based and game-based
methods operate as per Figure \ref{planBlocks}b for the case of a given,
fixed environment.   In these instances, novel or altered environments
require additional training to re-learn.  It is not clear that they
could ever really operate at the deliberative or global level for
general purpose navigation tasks as the underlying model complexity is
quite large.  None of the methods described operate as per Figure
\ref{planBlocks}c.  This paper is concerned with the closed-loop
consequences of traditional end-to-end ML pipelines while exploring
means to achieve a fused system with ML for the local/reactive planner
and a model-based approach for the global/deliberative planner.  

As foundational elements for our local/global fusion, we utilize
traditional methods that work at the local and global levels,
such as dynamic window approach (DWA) \cite{dynamicwindow}, 
elastic bands (EB) \cite{EB}, and timed elastic bands (TEB) \cite{TEB},
and their existing ROS implementations. 
Thus, the investigation does not seek to propose new planning strategies,
nor new learning architectures, but rather it seeks to investigate how to
effectively integrate ML-based approaches into existing navigation
pipelines.    The remainder of this paper covers the deep learning
strategy, the planning integration, and test results from simulation and
real-world operation.

\section{Deep Learning}

The AlexNet \cite{alexnet} deep convolutional neural network serves as the
foundation for our model. To facilitate learning from few training
samples, pre-training of the network involves using the ImageNet
dataset \cite{deng2009imagenet} and its associated training task.  
We replace the final layer of the network with a shallow, single-layer learner (described in \ref{sec:mappings}).
Several learning tasks are explored, including the traditional
approaches from the literature: control action regression (continuous
output), and control action classification (discrete output). Prior to
explaining the different ouput spaces, we first describe the common
aspects of the training methodology. 

\subsection{Generating Data Samples}
As is rapidly becoming the standard, we train with simulated data
\cite{SaLe_RSS[2017],PfEtAl_ICRA[2017],MiSaNg_ICML[2005],MiEtAl_ICLR[2017]}.
We use the Gazebo simulator to model our robot and a depth sensor in
various scenarios.

\vspace*{0.05in}
\noindent
{\bf Capturing scenes:} Three (roughly cylindrical) trash cans are randomly (uniformly) placed
in a rectangular region in front of the robot. The region extends from 1
to 5 meters in front of the robot and 3 meters to either side. Not all
of the region can be seen by the depth sensor, resulting in some
scenarios with fewer than 3 trash cans in them. A depth image is captured
by the depth sensor then decimated by 4, resulting in image samples with
dimensions 140x120.

\begin{figure}[t]
  \centering
  \vspace{.15cm}
  \includegraphics[width=0.75\columnwidth]{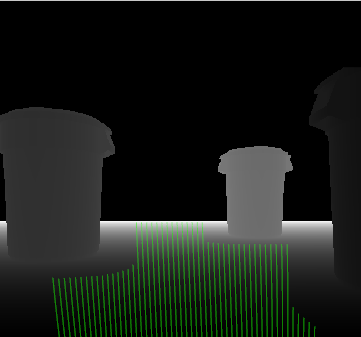}
  \caption{Training Sample: Depth image with non-colliding trajectory
    poses overlaid in green.  \label{fig:training}}
  \vspace*{-0.20in}
\end{figure}

\vspace*{0.05in}
\noindent
{\bf Generating Trajectories:} Using the trajectory generation and 
collision-checking approaches of \cite{smith2017pips}, we evenly sample 51 
robot departure angles in the range $[-0.4, .4]$ radians.  It involves
converting the departure angle into a feasible trajectory, generating a
sequence of poses along the trajectory, then checking each pose for collisions. 
The non-colliding poses per departure angle reveal how far the robot can
safely travel for each departure angle (see Figure \ref{fig:training}). 
This approach will implicitly encode the geometry of the robot relative
to the geometry of the world, as the collision checking employs a solid
model of the robot.  A robot with different geometry, motion
kinematics/dynamics, and dynamic constraints  would need this training
phase to incorporate those models.  Our model was the TurtleBot 2 mobile
robot whose constraints are obtained from the manufacturer's datasheet.

\vspace*{0.05in}
\noindent
{\bf Retaining Final Pose:} 
We retain the last non-colliding pose of each trajectory tested.  The
pose information is later converted into the appropriate output signal
based on the desired ouput space. The task metric employed for a given
learning scenario dictates how to use the stored final pose
information.  

\vspace*{0.05in}
\noindent
{\bf Generating Dataset Statistics:} 
We calculate the the per-pixel means and standard deviations for the
dataset and save them with the rest of the data. The statistics are used
to normalize samples for training and inference.

\subsection{Learning the Model}

Implement and training of the model is done with Tensorflow; in
particular, using the Adam optimizer with a learning rate of .001. 
Training is performed with a batch size of 600 on an NVidia 1070GTX GPU.
Training images are uniformly sampled  from dataset. After each epoch, the model is
run against a separate testing set.
When a sample is selected either for training or for testing, the image
is normalized and resized to match the expected dimensions of AlexNnet
(227x227). The result is then copied across 3 channels (the original
network was intended for RGB images).

For classification-based output spaces, the loss function is cross
entropy with logits.    For regression-based output spaces, the loss
function is the $L_2$-norm.  Training ends when the loss levels off;
this occurs within 500 iterations ($\sim$35 epochs).

\subsection{Input/Output Mappings}
\label{sec:mappings}

The deep network model is intended to inform the trajectory selection or
search process by using depth images as inputs.  In this work, we use
the same family of trajectories as in our prior work
\cite{smith2017pips}. These trajectories are parameterized by the
initial robot velocity and a departure angle.  As a simplification, we
assume a constant forward velocity. Since these trajectories only
involve strong rotations for brief periods,  it is a reasonable
assumption.

We implement three versions with different output spaces.
The first two represent
implementations reported in the literature.
In all of the versions, the intermediate label representation is the
Euclidean distance from the body frame origin to the end pose of each
departure angle (relative to the body frame). 

\subsubsection{Regression}
In this version, the last layer of the AlexNet model is replaced with a
1024x1 fully connected layer with no ReLU.  The output of this model is
a departure angle.  The correct output is the departure angle
corresponding to the greatest distance. In one version, the regressor was
goal-agnostic, in another it is goal-informed.

\subsubsection{Best Angle Classifier}
In this version, the last layer of the AlexNet model is replaced with a 1024x51 fully connected layer with ReLU.
The 51 output values represent the confidences of 
the corresponding departure angle being the correct one.
The winning class is the departure angle with the greatest distance.  In
this strategy, there is competition amongst the classifiers.

\subsubsection{Collision-free Angles Classifier}

In this version, the last layer of the AlexNet model is replaced with a
1024x102 fully connected layer with ReLU, with the output reshaped to
51x2.  After applying softmax to the last dimension, each 1x2 pair
represents the positive and negative confidences of a departure angle
being non-colliding.  A departure angle is classifed as non-colliding if
it's distance is greater than or equal to 4.0m, the approximate
effective range of a Kinect.  In contrast to the \textit{Best Angle
Classifier}, this version yields a family of independent (non-competing)
51 binary classifiers.

The last classifier is motivated by our experiences with the first three
models trained (the two regression and {\em best angles} classifier),
which did not have strong performance (see Section \ref{ExpEval}).
Evaluating trajectories for collisions is computationally costly,
therefore we seek to reduce the number of trajectories that we need to
evaluate in order to find a satisfactory one.  The purpose of the model
is to learn a mapping between depth images and departure angles. 
The model should not dictate which departure angle to use, however,
as that would leave no room for higher level planning systems to
operate. Rather, it should indicate which departure angles are likely to
yield good (i.e. non-colliding) trajectories.

\section{Planning Framework}

At every local path-planning invocation, the depth image is provided to
the trajectory generator to generate an output.
For the ML-navigation strategies that lead to a single outcome, the
high-level planner plays little to no role.  For all other strategies,
the samples get passed to a trajectory evaluator that then selects the
locally optimal choice. We incorporate Dynamic Window Approach, Elastic Band, 
and Timed Elastic Band local planner options within a global planning
framework using the ROS \textbf{move\_base} package. These planners
served as the baseline planner approaches for the comparison with
ML-integrated global/local planners. 

The ROS path planning and navigation stack supports several cost
functions to guide the navigation.  With respect to the current
position and orientation of the robot, the scoring functions included in
our navigation objective function are the following:
\begin{enumerate}
  \item Goal Heading: robot directed relative to goal direction;
  \item Path Heading: path keeps the robot nose on the nose
  path;\footnote{The ``nose'' is a virtual point ahead of the robot that
  exploits differential flatness to simplify the feedback control
  signal. The planned path also has such a point. It implicitly encodes
  the path tangent.}
  \item Path Distance: path proximity to to global path;
  \item Goal Distance: path leads robot closer to goal; and
  \item Obstacle Cost: path avoids obstacles.
\end{enumerate}
They score the sampled trajectory according to the preferences stated
above.  The first two score the orientation component of the path, the
second two score the position, and the last embodies obstacle avoidance.
Once the trajectory that the robot should follow is selected, the
trajectory is then converted into velocity commands and executed in the
low-level trajectory controller.
The navigation system follows the path until the next planning stage. 
For laser scan-based local planners that store obstacles in a
Cartesian occupancy grid, re-planning is triggered every 1 second.  The
laser scan data is simulated from a single horizontal line from
the depth image. Cartesian grid planners with this laser scan data will
inherit any field-of-view limitations associated to the depth image.
The perception space local planners \cite{smith2017pips} re-plan when the
robot achieves 60\% of the planned trajectory from the last planning
stage or if it detects a collision along the predicted path (occurs with dynamic
obstacles).

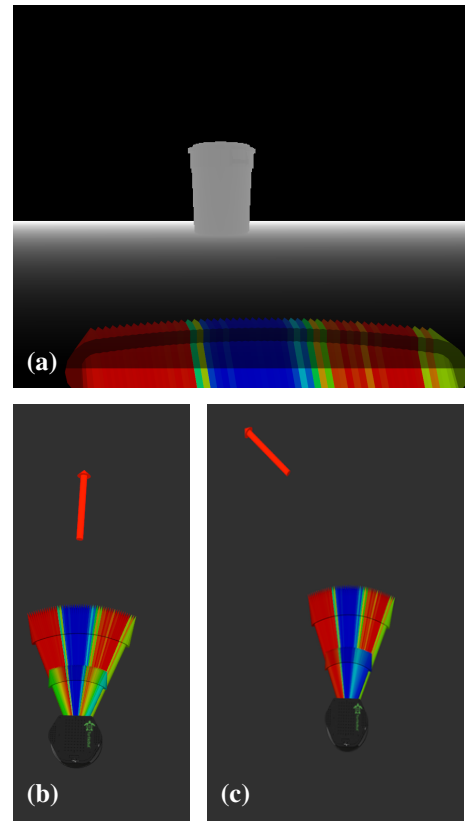
\begin{figure}[t]
  \centering
  \vspace{.05cm}
  \begin{tikzpicture}[outer sep=0pt]
    \node[anchor=south west] (MI) at (0in,0in)
      {\includegraphics[height=2in]{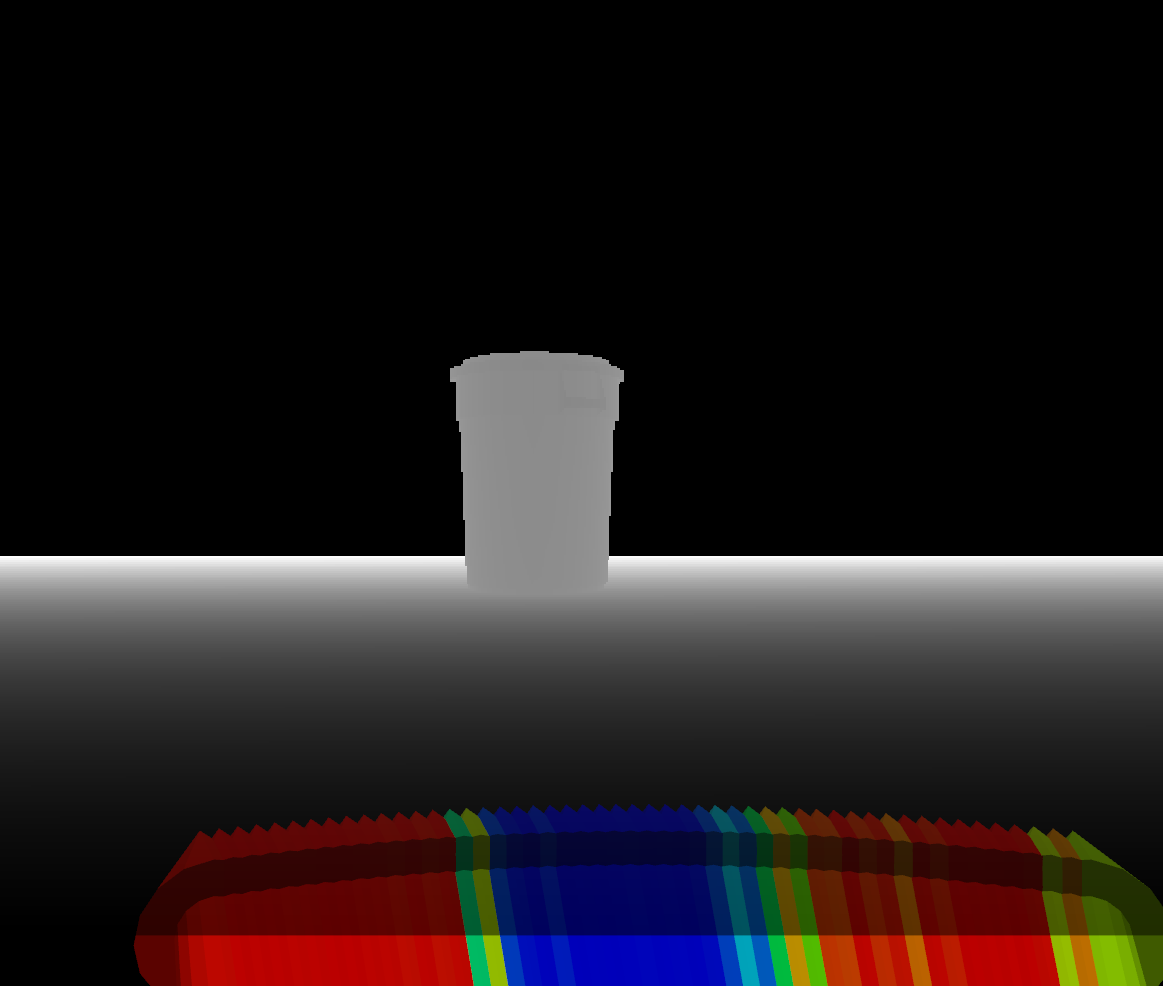}};
    \node[anchor=north west] (GB1) at (MI.south west)
      {\includegraphics[height=2.175in, trim={15cm 10cm 15cm 5cm},clip]{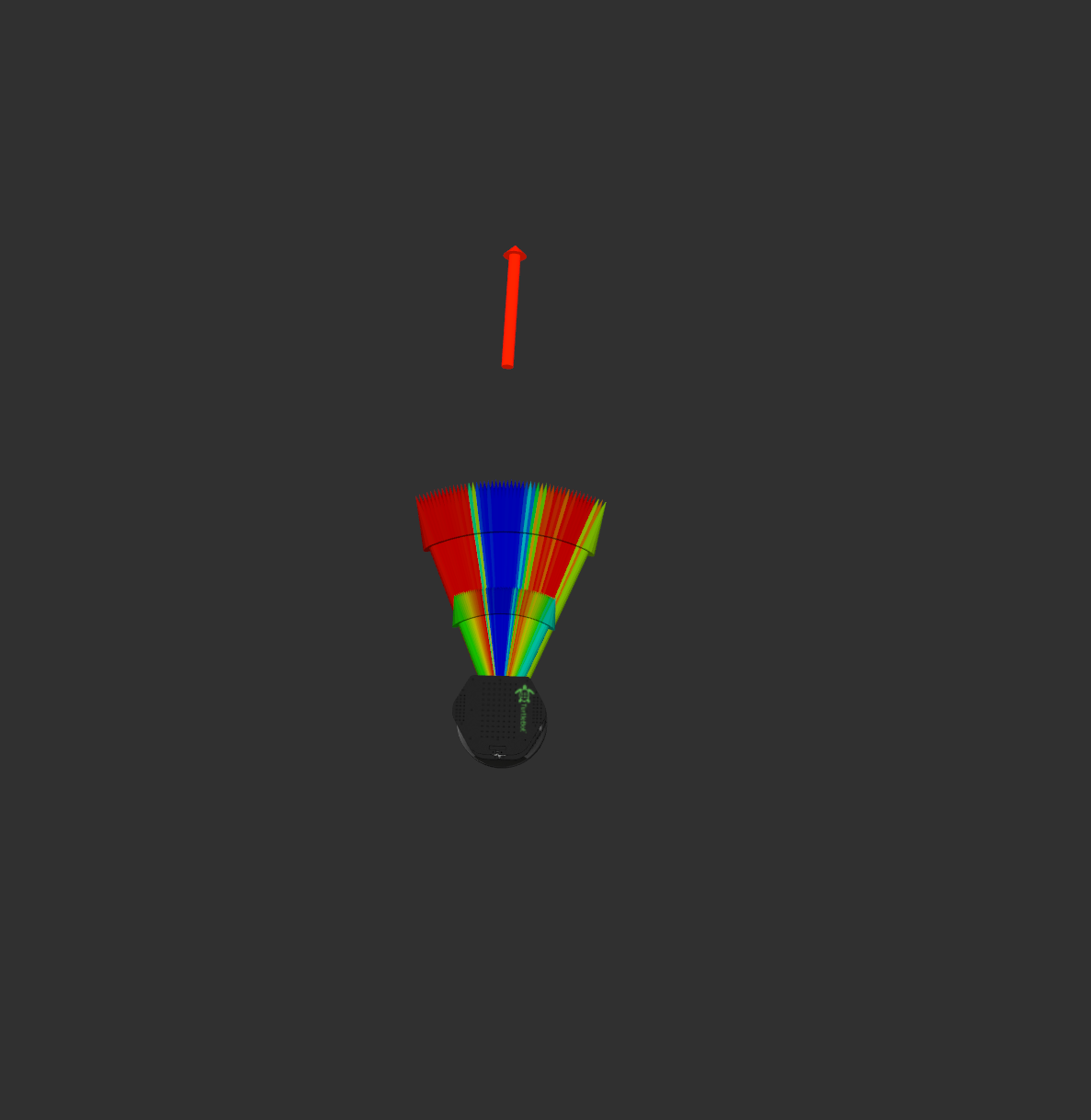}};
    \node[anchor=south west] (GB2) at (GB1.south east)
      {\includegraphics[height=2.175in, trim={3cm 1cm 5cm 1cm},clip]{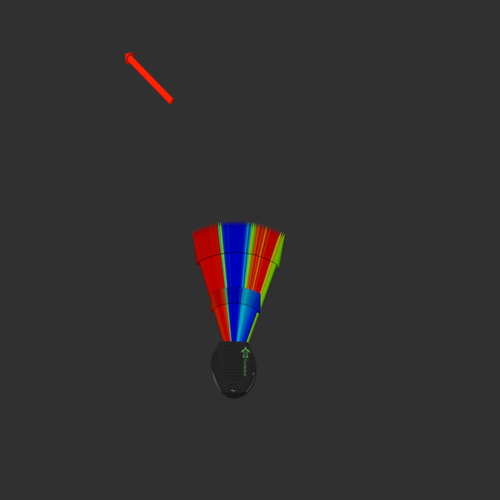}};
    \node[anchor=south west, xshift=5pt, yshift=5pt,color=white] at (MI.south west) {\bf (a)};
    \node[anchor=south west, xshift=5pt, yshift=5pt,color=white] at (GB1.south west) {\bf (b)};
    \node[anchor=south west, xshift=5pt, yshift=5pt,color=white] at (GB2.south west) {\bf (c)};
  \end{tikzpicture}
    \caption{Sample scenario with a barrel straight ahead. (a) Depth image 
      shown with color coded rays indicating the classifier confidence
      of each angle. Red denotes highest confidence and blue lowest.
      (b) Depiction of goal influence on confidence.  The longer vectors
      are the original confidences, while the shorter vectors of the
      inner wedge of rays has the goal-influenced confidences. For a
      straight ahead goal the left and right splits are equally high in
      the directions that minimize path length. (c) Depicts the same for
      a goal to the left.  Now the confidences are biased towards the
      goal and there is no left/right split for the goal-influenced
      confidences.
      \label{fig:nogoal}
      \label{weighted}
      }
  \vspace*{-0.30in}
\end{figure}

Unlike traditional learning based approaches that are usually trained
without awareness of the goal state, our system implemented two
additional methods on the top of trajectory selection algorithm to find
both locally and globally optimal candidates. One is a ``to-goal''
approach where the departure angle that points most directly to the
goal is added in addition to the top \textit{k} departure angles from
the model. Another is ``Gaussian-based goal bias'' approach which biases
trajectory selection based on the relative position of the goal. 
It weights the raw classificaton confidence scores by a Gaussian prior 
centered on the departure
angle that would point directly to the goal. %
In Figure \ref{weighted}, long
arrows represent unbiased trajectory candidates while the shorter arrows 
indicate the weighted values. 

\section{Experimental Evaluation}
\label{ExpEval}
To evaluate the quality of the overall ML-enhanced navigation pipelines, we
run the vision-based navigation system in Gazebo-simulated environments
and on an actual mobile robot (Turtlebot).  Simulation serves to create
highly repeatable and consistent test conditions for evaluating a variety
of planning implementations.

In what follows, we do not report the results for the regression-based
approaches.  Both the goal-agnostic and the goal-informed regression
approaches perform poorly and can not navigate without collision,
in contrast to \cite{TaLiLi_IJARS[2017]} who report success at a
regression approach, as well as \cite{PfEtAl_ICRA[2017]} whose method is
goal-informed. We believe differences in training sets may be responsible for
the differences in performance.  
Our training scenes involve randomly placed obstacles in the field of
view and simulate only one instance in time, whereas those from
\cite{TaLiLi_IJARS[2017]} appear to be mostly corridor images that arise
from actually executing a path. Our learnt model would have no notion of
a corridor in its internal representation. The model in
\cite{PfEtAl_ICRA[2017]} is trained in a sparser environment with
large obstacles using navigation commands from a global planner.
Topologically distinct trajectory splits would not be as frequently
observed in those models.

\subsection{Barrel Forest}
The first battery of navigation tests in a simulated environment
involves positioning barrels in random locations within a 10x6$m^2$
world where the robot's goal is to travel 8 meters forward.  the number
of barrels randomly spawned is set to 3, 5, and 7 across three sets of
trials.  For each quantity of barrels, 50 scenes are created.  For every
instantiated scene, we initialize the navigation pipeline with each of the 
following local planners:
\begin{enumerate}
  \item Dynamic Window Approach %
  \item Elastic Band  %
  \item Timed Elastic Band  %
  \item Learning PiPS + To-Goal  %
  \item Learning PiPS + Goal-Biased %
  \item Learning Cartesian + To-Goal 
  \item Learning Cartesian + Goal-biased 
  \item Learning Naive + Goal-Biased %
\end{enumerate}
where PiPS is the perception space planner \cite{smith2017pips}.  
The last planner tested is the traditional end-to-end planner with a
discrete steering space and a winner-takes-all multi-class classifier
(sometimes called the ``naive'' planner as it represents a typical first
pass solution to the end-to-end learning problem, after regression).
However, it also includes goal biasing.
The objective of each
task is to simply move from one end of the world to the other end
without crashing. 

The first three approaches described above use traditional exhaustive
trajectory search, creating 200 trajectory candidates at each
replanning phase.  
The learning-based approaches generate trajectories using 
the five departure angles with the
highest confidences. For the Naive approach, the
trajectory with the highest confidence is executed without scene
analysis (i.e. no collision checking). 
In order to test the
original quality of the trajectory selection strategies, we disable the
recovery behavior performed by the global planner when the local planner
reports being stuck.

\subsubsection{Controller Comparison}
\begin{figure}[t]
  \centering
  \includegraphics[width=9.5cm,trim={3.5cm 0cm 0cm 1cm},clip]{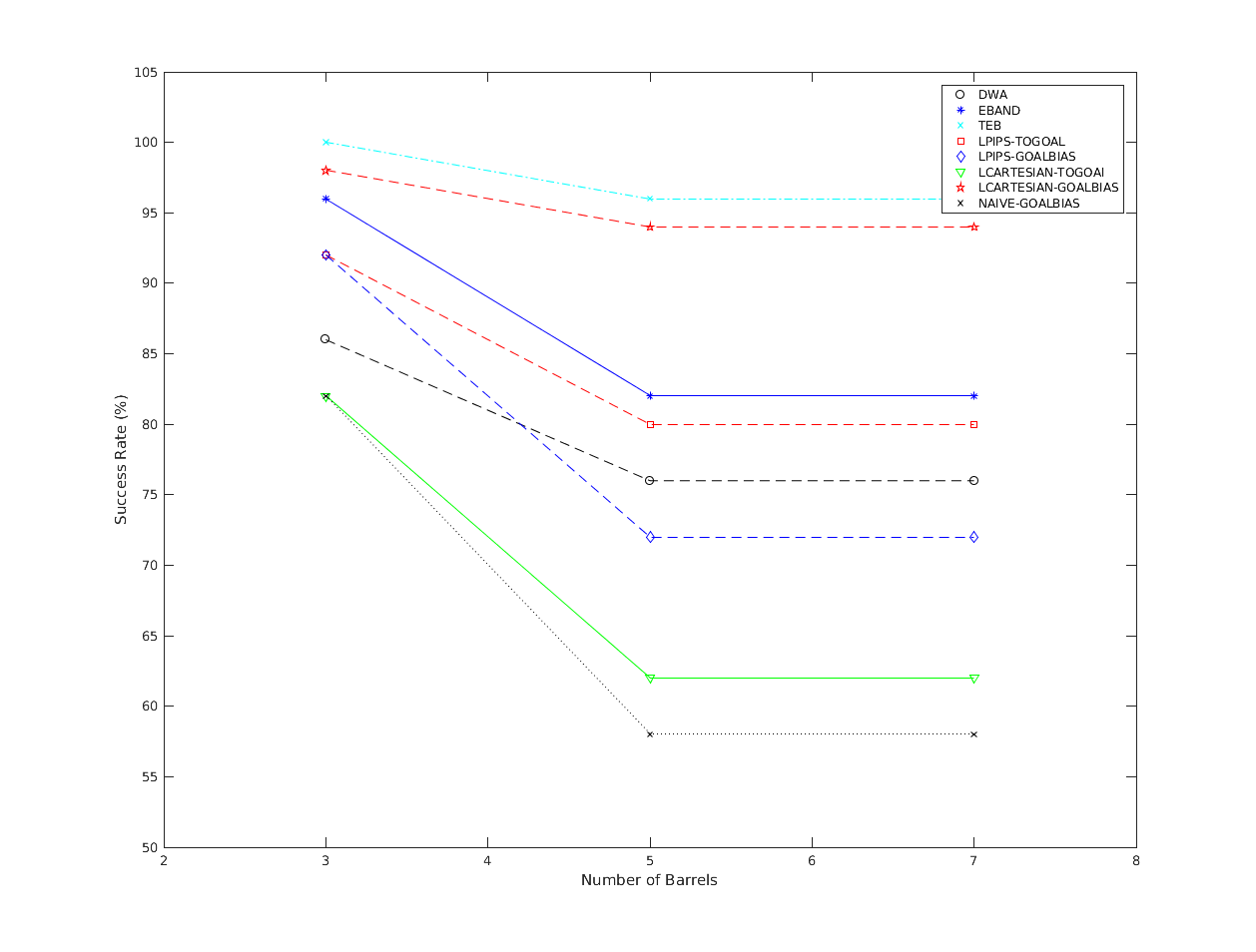}
  \caption{Barrel Forest Success Rate}
  \label{exp1}
  \vspace*{-0.15in}
\end{figure}

Figure \ref{exp1} graphs empirical results for the Barrel forest
navigation task. Recall that performance should not be perfect due to
(1) disabling of the local-minima recovery behavior and (2)
field-of-view limitations that mean the robot might turn into an unseen
object.
In this experiment, the TEB local planner consistently has the highest success rate
while the Naive controller has the lowest.
However, the experiment shows that
path-planning using only five ML-predicted trajectories per planning stage
can achieve a success rate close to traditional exhaustive trajectory
search-based planners.  The ML methods sample only $\frac{1}{40}$ as many trajectories
as the exhaustive searchers. 

The ML Cartesian planner with Gaussian goal biasing
performs only 2\% worse than TEB and outperforms EB and DWA. Learning PiPS with
Gaussian goal biasing outperformed DWA.  The difference in performance
is that the Cartesian planners incorporate some memory in the local
cost-map, whereas PiPS has no memory.  It runs a higher risk of turning
back into objects that leave the field of view when being passed.
Adding memory should improve performance.

\subsubsection{Trajectory number comparison}
Next, we test the impact of varying the number of trajectories generated by the ML
planners.  The number
of barrels is fixed to 5 while the number of trajectories sampled is set to
${2,3,5,7}$ across four trial sets.  There are 50 trial scenes with
randomly spawned barrels.  To increase the difficulty of the scenario,
the barrel spawn zone is made smaller, thereby increasing the density
of the ``barrel forest.''
We use the Learning Cartesian planner with Gaussian Goal-biasing since it has the
the best performance of the ML approaches we are testing. As in the previous
experiment, we disable recovery behaviors.
From Table \ref{exp2_table}, the experimental results show that the
number of trajectory candidates provided by the learning-based model does not
have a significant impact on sucess rate as long as more than three candidates are
provided into the evaluation. 

\begin{table}[t]
 \centering
  \caption{Success Rate vs Trajectory Candidate Quantities}
  \label{exp2_table}
 \begin{tabular}{||c || c c ||} 
 \hline
 \# candidate trajectories & Success Count & Rate \\ [0.5ex] 
 \hline
 2 & 29 & 58\% \\
 \hline
 3 & 36 & 72\% \\
 \hline
 5 & 38 & 76\% \\
 \hline
 7 & 39 & 78\% \\
 \hline
\end{tabular}
\vspace*{-0.1in}
\end{table}

We examine the failure cases to determine their sources.
In some of these cases, %
the classifier failed to return valid trajectories when an exhaustive search
would have succeeded.
These cases could be resolved by one of following approaches:
switching to an exhaustive search when the ML approach fails, or
replanning the global path using the information gathered from
local planners.

Exhaustive searching should be avoided if possible, but can always be tried as a fallback approach.
In order to test the second approach,
we rerun the 11 failure cases from the scenario with 7 candidate trajectories
with the recovery behavior enabled.
In 6 out of 11 times, a new global plan 
is found that successfully reaches the goal.   The remaining 5 cases
fail due to the recovery behavior being tailored to laser scan models.
The robot turns 360 degrees and returns to face the same scene.
A more standard laser scanner would most likely provide better
information than the depth limited kinect scanner.  One option may be to
create a new recovery behavior for visual
navigation strategies.

\subsection{Sector World}
\begin{figure}[t]
  \centering
  \vspace{.15cm}
  \hfill
  \includegraphics[width=0.49\columnwidth]{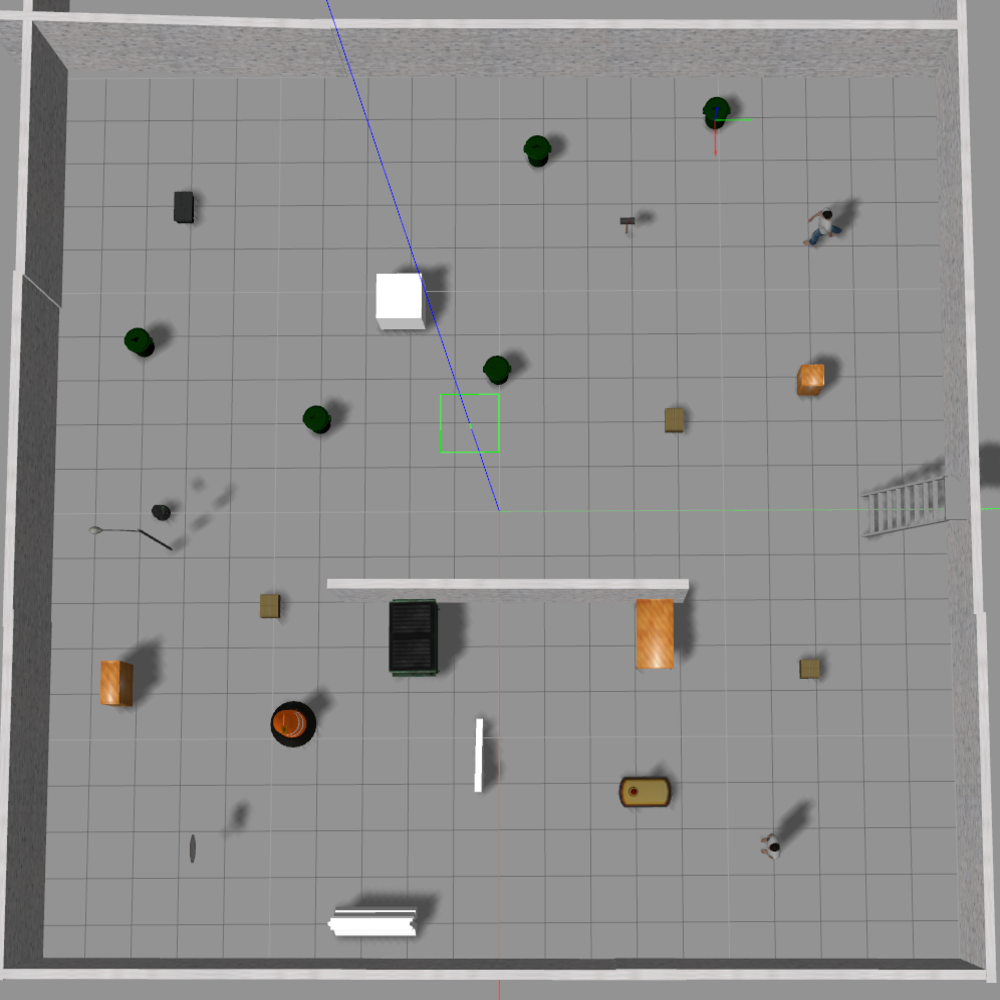}
  \hfill
  \includegraphics[width=0.49\columnwidth]{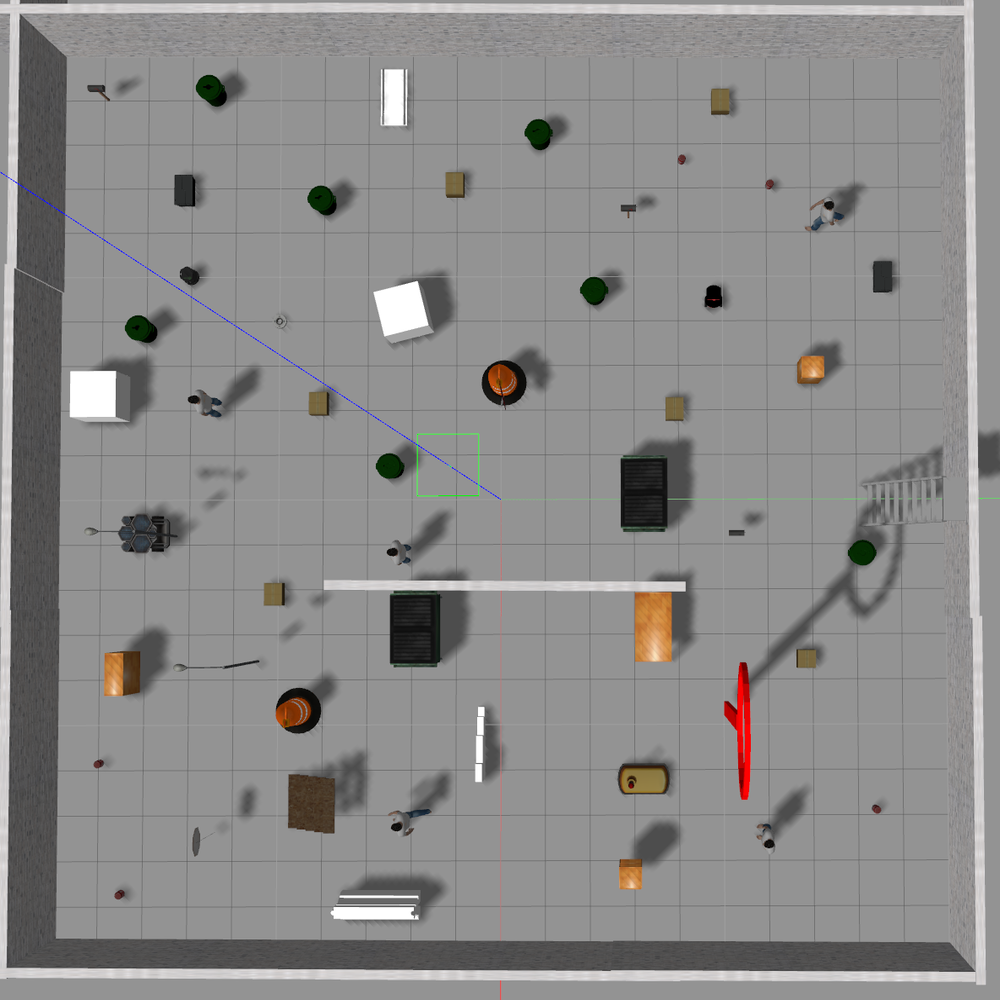}
  \hfill
  \caption{Sparse (left) and dense (right) sector worlds.
  \label{fig:sector}}
  \vspace*{-0.15in}
\end{figure}

We also test our approach in environments that are divided into sectors
and populated with an assortment
of objects.
The two test environments  are depicted in
Figure \ref{fig:sector}, one sparsely populated and one more densely
populated with more diverse object categories (recall that only garbage
cans were in the training set).  
The object positioning remains static while the start and goal sectors
are randomly selected for each test. The objective of this experiment is
to navigate from the start zone and travel to the goal sector without
crashing. 
In this experiment we compare the best PiPS approach (Learning PiPS + To Goal) 
to the best traditional approach (TEB).
For both worlds, we test 35 navigation sector-pairs, again with the
recovery behavior disabled.  The results are in Table \ref{exp3_table}. 

\begin{table*}[t]
\centering
\vspace{.1cm}
\caption{Experimental results for Sector World.}
\label{exp3_table}
\begin{tabular}{|l|c|c|c|c|}
\hline
   & \multicolumn{2}{c|}{World A (Sparse)} & 
     \multicolumn{2}{l|}{World B (Dense)} \\ \hline
   & Success Rate (\%) & Avg. Time (s) & 
     Success Rate (\%) & Avg. Time (s)\\ 
   \hline
LPIPS-ToGoal & 93.33            & 57.55                 & 70               & 66.02                 \\ \hline
TEB          & 86.56            & 55.8                  & 90               & 63.41                 \\ \hline
\end{tabular}
\end{table*}

The ML-based trajectory generation strategy outperforms the exhaustive
trajectory TEB local planner for the easy sector world.  In denser environment,
 the success rate decreased 23\% for the ML planner 
and increased about 3.5\% for the TEB planner.  
The less than perfect
completion rate for both approaches reinforces the well-accepted observation that
local planners need augmentation by global planners with recovery
behaviors, as well as the need to consider
the full navigation pipeline when creating and evaluating ML-based
image-to-decision navigation methods.

\subsection{Real World Implementation}

\begin{figure*}[!tbp]
  \centering
  \includegraphics[width=0.24\textwidth]{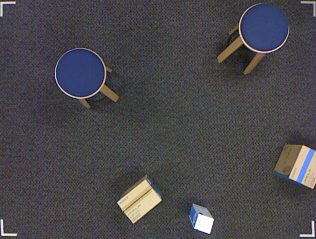}
  \includegraphics[width=0.24\textwidth]{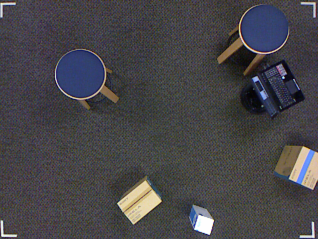}
  \includegraphics[width=0.24\textwidth]{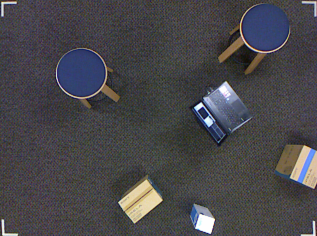} \\
  \vspace*{0.05in}
  \includegraphics[width=0.24\textwidth]{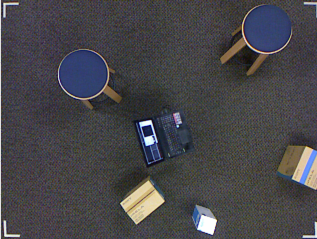}
  \includegraphics[width=0.24\textwidth]{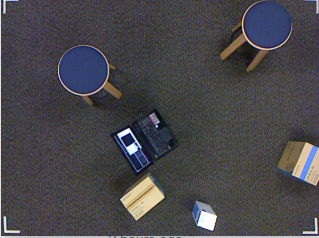}
  \includegraphics[width=0.24\textwidth]{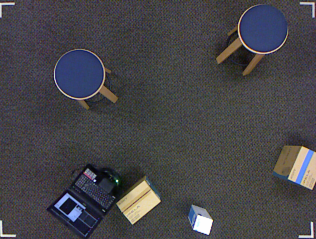}
  \caption{Overhead snapshots of Turtlebot navigating an obstacle course (chronologically: from top left to top right, then bottom left to bottom right).
    \label{TBot}}
  \vspace*{-0.2in}
\end{figure*}

Lastly, we implemented our ML-based planner module on a Turtlebot 2
equipped with a laptop.
Given a goal and a known starting position, the robot is be able to
navigate through obstacles with only 5 ML-generated trajectory
candidates.   In 13 of the 15 tests, navigation from start to goal is
successful.  For 2 of the 15 cases, the Turtlebot collides with objects
located just outside the field of view.  Figure \ref{TBot} consists of
overhead views of one navigation scenario.  The Turtlebot starts off
screen at the upper-right (just past the barely visible white angle
marker) and is directed to travel to another off-screen location at the
lower-left of the image, offscreen.  The start and final positions are
chosen such that following the straight line path results in
collisions.   In the last image one can see how closely the PiPS planner
tends to get to obstacles when passing them.

\section{Conclusion}
This paper explores the behavior
of typical end-to-end machine learning based navigation strategies 
under controlled simulation conditions, as
well as proposes a revised training and trajectory generation method
based on the outcomes of the strategies.  While demonstrating success at
collision-free wandering in natural, real-world environments, the
traditional ML algorithms perform poorly when faced with a goal-directed
navigation task.  By modifying the input/ouput mapping, adjusting the
training data provided and its evaluation criteria, and incorporating
well established goal-directed modifications to the planning pipeline, a
deep learning based modification to existing navigation stacks is
described.  In controlled conditions, the proposed deep learning
navigation strategy is shown to operate as effectively as commonly used
methods in the robotics community. %(sometimes better, sometimes worse).
%Furthermore, the compute time associated with finding the trajectory was
%significanlty reduced (factor of 40x).  In doing so, the speed-ups
%enabled more complex 3D collision checking to be performed.
Furthermore, the number of trajectories needed to find a path
is significantly reduced.

Further work is needed to resolve the issues that arise from the
limited field of view associated to dense visual sensors. One solution
would be to incorporate additional memory within the local planner,
whereas another would be to modify the model and training procedure to
consider the time history of decisions in order to implicitly learn to
create a buffer away from obstacles when passing them.

\bibliographystyle{IEEEtran}
\bibliography{regular,refs} 

\end{document}